\documentclass[11pt]{article}
\usepackage[final]{acl}
\usepackage{times}
\usepackage{latexsym}

\usepackage{natbib}
\usepackage[utf8]{inputenc} % allow utf-8 input
\usepackage[T1]{fontenc}    % use 8-bit T1 fonts
\usepackage{hyperref}       % hyperlinks
\usepackage{url}            % simple URL typesetting
\usepackage{booktabs}       % professional-quality tables
\usepackage{amsfonts}       % blackboard math symbols
\usepackage{nicefrac}       % compact symbols for 1/2, etc.
\usepackage{microtype}      % microtypography
\usepackage{xcolor}         % colors
\usepackage{amsmath}
\usepackage{multirow}
\usepackage[capitalise]{cleveref}
\usepackage{graphicx}
\newif\ifshowcomments
\showcommentsfalse % Set \showcommentstrue to display coauthor comments.

\title{VoiceLongMemEval: Do Assistants Remember How You Sounded?%
\thanks{Dataset:~\url{https://huggingface.co/datasets/ramitpahwa13/voice-longmemeval}.}}

\author{%
  Ramit Pahwa \\
  \texttt{ramitpahwa@rivianvw.tech}
  \AND
  Parivesh Priye \\
  \texttt{pariveshpriye@rivianvw.tech}
  \AND
  Apoorva Beedu \\
  \texttt{apoorvabeedu@rivianvw.tech}%
}
\begin{document}

\maketitle

\begin{abstract}
% Vocal delivery can change how a user's words should be interpreted. Yet a correct answer about an earlier conversation may not come from transcript cues alone. We introduce VoiceLongMemEval (VLME), a benchmark that tests how delivery cues affect answers about earlier conversations. The Benchmark has ~500 curated scenario [Need to add 1-2 libes here]. Our main evaluation holds each history and question fixed and compares transcripts with and without textual descriptions of vocal delivery. Across eight models in roughly 10--15k-token contexts, descriptions improve accuracy by 8.9--38.3 percentage points; On derived indirect questions from our original dataset. Our result shows why prompting gains alone cannot establish that an assistant used delivery cues. Code and data will be released upon acceptance.
Vocal delivery can change how a user's words should be interpreted. Yet a correct answer about an earlier conversation may come from transcript cues alone. We introduce VoiceLongMemEval (VLME), a benchmark with 316 questions across seven tasks: 202 direct questions, and 114 indirect questions. Conversations are LLM-authored, and every question undergoes transcript-only adversarial checks. Our main evaluation holds each history and question fixed and compares transcripts with and without textual descriptions of vocal delivery. Across eight models on the full benchmark in roughly 10--15k-token contexts, descriptions improve accuracy by 12.9--41.7 percentage points; the highest three-seed mean accuracy is 56.1\% under the standard prompt. Full-benchmark ablations compare acoustic descriptions, explicit emotion labels, structured annotations, and prompting.

\end{abstract}

\section{Introduction}
\label{sec:intro}

% ``That was an okay restaurant'' can sound like approval or disappointment. If a user later asks whether they enjoyed the meal, remembering the words may not be enough. A voice assistant needs to preserve the relevant delivery and connect it to the later question. The challenge extends beyond recognizing emotion in a clip: evidence may be scattered across sessions, preferences may depend on a previously expressed state, and a later utterance may change an earlier interpretation.

A user tells their assistant, in a flat voice trailing into a sigh,
``That was an okay restaurant''. 
Weeks and a hundred thousand tokens later, when the user asks, \emph{How did I feel about that restaurant the other day?}, everything needed to answer was present at encoding time, but only in the delivery. 
An assistant that transcribed it would answer, ``You found it okay!''. 
But only the model that would have attended to the emotion would have known the user did not really like the restaurant.
Oftentimes, the challenge extends beyond recognizing emotion in a clip.
More broadly, assistants must retain and update such paralinguistic evidence across long, multi-session histories yet most benchmarks still evaluate memory and perception in isolation.

% Conversational memory benchmarks such as LoCoMo and LongMemEval evaluate recall, updates, and reasoning over dialogue histories~\citep{maharana2024locomo,wu2025longmemeval}. Speech benchmarks test paralinguistic perception and contextual interpretation~\citep{cpbench2025,anand2026parallm}. VoiceMem also evaluates long-horizon paralinguistic memory through ChatMem-Bench~\citep{xie2026voicemem}. These advances make a more specific question useful: when a later answer depends on delivery, how much does a model gain from retaining that evidence rather than the transcript alone?
While long-term conversational memory~\citep{maharana2024locomo,wu2025longmemeval,jiang2025personamem,wu2026longmemeval}, and paralinguistic perception ~\citep{ao2024sdeval,yang2024airbench,cpbench2025} have both become measurable capabilities, the two areas have only been tested separately so far.
Their intersection, \textit{i.e} remembering how something was said, retaining it across sessions, updating any delivery shifts, and retrieving it against a specific question much later, is a vital cross-modal dimension that current benchmarks lack.
While recent work like VoiceMem~\citep{xie2026voicemem} addresses voice-grounded memory, our benchmark strictly isolates paralinguistic memory by requiring every every item to fail an adversarial transcript-only baseline.

% \apoorva{-------remove}
% VoiceLongMemEval (VLME) addresses this question with paired views of the same conversational history. The blind view contains the transcript; the descriptive view adds acoustic descriptions of the user's delivery. We author a 202-question core across six task types and use an adversary to reject transcript-solvable cases. The final audit finds no correct transcript-only answers among its 175 non-abstention items for the designated adversary. This is an empirical filter against shortcuts, not a guarantee that every model or prompt will fail. Another 321 questions reuse the evidence to study more explicit and more indirect ways of asking about it; these extensions are not independently adversarially filtered.

% Our contribution is a controlled benchmark and an analysis of what its controls reveal. Across eight models, delivery descriptions improve core accuracy by 8.9--38.3 percentage points. Ablations show how the benefit changes when models receive acoustic descriptions, structured tags, or explicit emotion labels. Prompting helps transcript-only models much more on the derived questions than on the core, exposing a distinction between answers inferred from dialogue and answers supported by retained delivery. 
% \apoorva{this -------}

To address this gap, we introduce \textbf{VoiceLongMemEval (VLME)}, a paralinguistic memory benchmark that augments conversation transcripts with acoustic descriptions of user delivery.
% item embeds a paralinguistic needle in a haystack of conversational sessions: the answer is recoverable only from the delivery metadata, never from the words alone, enforced by a three-part adversarial gate described in \Cref{sec:method}.
Our contributions are:
 % \item \textbf{VoiceLongMemEval (VLME) Benchmark:} 523 adversarially validated items, spanning six question types that factor paralinguistic memory into affect recall, affective preference, affect update, cross-session affect, temporal-affective reasoning, and prosody-disambiguated interpretation, each with abstention variants.
 % \item \textbf{Systematic Analysis of the Affect Gap:} We evaluate eight models (three proprietary, five open-weight), revealing a consistent affect gap of +0.09 to +0.38 across all systems (all p < 0.001). Through fine-grained component ablations and a five-tier phrasing spectrum, we show that emotion tags provide the strongest signal, chain-of-thought reasoning cannot substitute for missing context, and controlled counterfactuals isolate a net $+0.067$ accuracy gain driven strictly by metadata content.
\textbf{\emph{(i)} VoiceLongMemEval (VLME) Benchmark} contains 316 questions: 202 core questions spanning six task types and 114 indirectly phrased questions for evaluating performance across phrasing variants. Every question in this final set was individually checked using a transcript-only adversary and human in the loop verification. This adversary is a model that attempts to answer questions from the conversation text alone, without access to audio or annotations describing emotion, prosody, or vocal delivery. Each question is accompanied with 100k token conversation history.
 \textbf{\emph{(ii)} Systematic Analysis:} Across models, delivery descriptions improve core accuracy by 12.9--41.7 percentage points. We compare descriptions, structured tags, emotion labels, and prompting to identify which inputs help and whether answers can also be inferred from transcripts alone.

\section{Related Work}

\textbf{Long-term conversational memory.}
Recent benchmarks evaluate multi-session retention~\citep{xu2021beyond,maharana2024locomo}, scalable retrieval~\citep{wu2025longmemeval,wu2026longmemeval}, personalization~\citep{du2024perltqa,tan2025membench,kim2024dialsim}, and implicit preferences~\citep{zhao2025prefeval,jiang2025personamem}. 
A-MBER~\citep{wen2026amber} tracks emotional states across multi-session dialogues, but its evidence is purely lexical. 
However, these existing evaluations test memory of \emph{what} was said. 
VLME extends this paradigm by requiring the paralinguistic feature \textit{i.e. }\emph{how} it was said, and models ability to remember to answer the query.

\textbf{Paralinguistic understanding in speech-LLMs.}
Audio-language benchmarks probe acoustic attributes, prosody, and emotion~\citep{huang2024dynamicsuperb,yang2024airbench,wang2025audiobench,sakshi2025mmau,wang2025mmsu}, contextual paralinguistics~\citep{ao2024sdeval,cpbench2025}, and speech interaction~\citep{jiang2025s2sarena,chen2025paras2s}, expanding on classical affective datasets~\citep{busso2008iemocap,poria2019meld} and LLM successors~\citep{xu2024secap,lin2024paralinguistics}. 
However, these focus on single utterances.
ParA-LLM expands coverage across 22 acoustic attributes with ParA-Bench~\citep{anand2026parallm}, while VoxParadox~\citep{pang2026voxparadox} intentionally mismatches transcript claims and speaking styles, exposing that audio-LLMs heavily rely on language-implied shortcuts over acoustic evidence. These findings motivate both delivery-aware evaluation and controls for lexical shortcuts.
VLME applies those controls to recall, updates, and interpretation across conversational sessions. Its metadata track supplies delivery descriptions explicitly, separating their use in reasoning from the ability to extract them from speech.
While VoiceMem~\citep{xie2026voicemem} targets long-horizon acoustic memory, VLME isolates whether delivery is strictly necessary by enforcing an adversarial transcript-only check over histories up to $\sim$100k tokens with audios.

\textbf{Cascaded vs.\ audio-native pipelines.}
Cascaded architectures (ASR $\rightarrow$ LLM) discard acoustic nuances at the transcription boundary, whereas audio-native models~\citep{hurst2024gpt4o,chu2024qwen2audio,xu2025qwen25omni,defossez2024moshi,zeng2024glm4voice} ingest speech directly. While prior cascade comparisons focus on single-turn tasks~\citep{ao2024sdeval,cpbench2025,pahwa2026audio2tool}, VLME evaluates paralinguistic loss at the memory level, where unrecorded prosody in early sessions silently breaks recall weeks later.
\section{Benchmark Construction}
\label{sec:method}

% VLME contains 316 questions linked to 326 annotated evidence sessions: a 202-question core, and 141 indirect questions that ask about an action, preference, or stance without mentioning voice. The derived sets reuse evidence from the core: they share the same underlying ``haystack'' conversation and construct different question variants over it, rather than representing 321 independent conversations. Records follow the LongMemEval format~\citep{wu2025longmemeval}.
VLME contains 316 questions: a core 202-question split and 114 indirect questions split about actions, preferences, or stances without explicitly mentioning vocal delivery  spanning seven task types. All questions underwent transcript-only adversarial checks. Each item contains a question, reference answer, and timestamped conversation history with paralinguistic annotations, adapting the LongMemEval record format~\citep{wu2025longmemeval}.

\subsection{Annotations and Tasks}
\label{sec:schema}
\label{sec:questions}

Every user turn is annotated with an emotion label, a prosody tuple (rate, pitch, loudness, pauses, and emphasized words), vocal events, sarcasm and uncertainty flags, and a delivery description. Descriptions use acoustic stage directions such as ``quick and light, laughs mid-sentence'' and exclude emotion names or interpretations such as ``relieved.'' The main descriptive condition adds these descriptions to the transcript. In the reported core ablations, emotion-only input supplies explicit emotion labels, while tagged input supplies structured annotations (\Cref{tab:ablation316}). Evaluating these inputs separately lets us compare descriptions of how a user spoke with explicit labels of the emotion conveyed.

The benchmark covers seven tasks (\Cref{tab:types}); intent attribution appears only in the indirect set. Answerable questions are designed to require interpreting delivery alongside the words. The 202 direct questions include 27 abstention cases that ask about an emotional episode absent from the history. The 114 indirect questions ask about actions, preferences, stances, or intentions without explicitly mentioning vocal delivery. They reuse evidence sessions from the direct set. Their task distribution differs from the direct set, so aggregate comparisons reflect both phrasing and task composition.

\begin{table*}[t]
\centering
\small
\caption{ The VLME benchmark has 316 questions which are naturally phrased  over the same evidence sessions. In the examples, \emph{italics} mark what the words say and \textbf{bold} marks the delivery cues that change the reading.}
\label{tab:types}
\begin{tabular}{p{2.9cm}rrrp{3.9cm}p{5.2cm}}
\toprule
Task & Direct & Indirect & Total & What must be remembered & Example \\
\midrule
Affect recall & 48 & 51 & 99 & A state expressed in an earlier turn & Dropping pottery sounds \emph{practical} in the words, but \textbf{slow speech} and \textbf{sighing} suggest \textbf{disappointment}. \\
Affective preference & 27 & 24 & 51 & A preference tied to an expressed state & A \textbf{``read aloud''} rule applies when the user sounds \textbf{frustrated}. \\
Affect update & 31 & --- & 31 & A changed reading of repeated wording & \textbf{Brighter delivery} changes the earlier interpretation of \emph{concern about chapter four}. \\
Cross-session affect & 30 & --- & 30 & A pattern across sessions & Delivery shifts from \textbf{flat and sighing} to \textbf{brighter} during knee rehabilitation. \\
Temporal-affective & 29 & --- & 29 & The order of affective episodes & Recall \textbf{which news came first} and \textbf{how each was delivered}. \\
Prosody disambiguation & 37 & 9 & 46 & A reading resolved by delivery & \emph{``Simplifies the plan''} sounds \emph{accepting} in text but is spoken \textbf{slowly and flatly}. \\
Intent attribution & --- & 30 & 30 & Whether an action was chosen or forced & A new cancellation rule is \emph{stated plainly}, but \textbf{slow, quiet speech}, \textbf{long pauses} and a \textbf{sigh} make it read as \textbf{forced}, not planned. \\
\midrule
Total & 202 & 114 & 316 & & \\
\bottomrule
\end{tabular}
\end{table*}

\subsection{Conversation Histories}
\label{sec:authoring}

An LLM writes conversations of 4--12 turns, with at most 26 relevant turns per question. We keep wording neutral where possible and avoid questions that give away the intended emotion. Delivery can differ from the reaction an event typically suggests: good news need not sound happy.

We provide histories containing only the relevant conversations (up to roughly 1k tokens) and full histories with additional unrelated sessions (about 100k tokens). 

% The reported experiments use relevant conversations plus five unrelated sessions, yielding roughly 10--15k tokens (\Cref{sec:setup}). The 100k-token histories have not been evaluated.

\subsection{Transcript-only audit}
\label{sec:audit}

While creating the question, we ask the model to answer from the transcript alone. We revise or discard questions it answers correctly. We also remove wording that gives away the intended emotion and test the model with delivery descriptions. This second check measures how much the descriptions help; it does not decide which questions are kept.

The check used Qwen2.5-72B-Instruct~\citep{qwen2technicalreport}, which found no correct transcript-only answers to the non-abstention questions under the tested prompt. Later checks cover every question in the final 316-question set. Each finding is specific to its model and prompt; other models or prompts may still answer correctly. Further details appear in \Cref{app:construction}.

\subsection{Synthetic audio}
\label{sec:audio_synthesis}

The audio evaluation covers the 202 direct and 114 indirect questions, using relevant-turn windows and evidence-session clips, respectively. We synthesized clips primarily with Dia~\citep{dia2025}, using acted emotional speech from RAVDESS~\citep{livingstone2018ravdess} as reference prompts for the target emotion. Three human reviewers assessed all clips; Qwen3-TTS~\citep{qwen3tts2026} and Maya-1~\citep{maya1_model} supplemented generation where review identified poor audio quality. These are controlled synthetic realizations of the authored conversations. Synthesis and emotion controls are described in \Cref{app:audio_synthesis}; the audio ablation appears in \Cref{sec:audio_results}.
\section{Experimental Setup}
\label{sec:setup}

We evaluate three proprietary models and five open-weight models listed in table \ref{tab:main316}
Our main comparison holds the conversation and question fixed while changing the input: \emph{blind} contains the transcript alone; \emph{descriptive} adds acoustic descriptions of vocal delivery.
The difference in accuracy is the \emph{affect gap}.

Each test case embeds its evidence among $n_d=5$ randomly sampled distractor sessions, producing roughly 10k--15k tokens of context.
Models answer in free text, and an LLM judge applies a task-specific rubric against the gold answer.
We use three seeds for distractor selection and arrangement and report mean accuracy $\pm$ standard deviation.
For the main paired comparisons, we use bootstrap resampling~\citep{efron2000introduction} (10,000 iterations) and McNemar's test~\citep{mcnemar1947note}.
The main affect gaps are reported as significant at $p<0.001$.
% Single-seed ablations are identified separately.

% The 202-question core is the primary evaluation.
% The Nuanced 181Q and Indirect 140Q sets test questions with explicit references to delivery and open-ended questions without such references, respectively.
% They were generated separately and did not undergo the core's adversarial gates; comparisons across these sets therefore reflect both question phrasing and item composition.
% Audio experiments use Indirect~v2 in an evidence-only setting.

\section{Results}
\label{sec:results}

\subsection{The Affect Gap on the Core Benchmark}
\label{sec:gap}

\begin{table*}[t]
\caption{Accuracy on the full 316Q release (202 original + 114 indirect; $n_d=5$, three seeds; mean $\pm$ standard deviation). The affect gap is $\Delta = \mathrm{descriptive}-\mathrm{blind}$.}
\label{tab:main316}
\centering
\small
\begin{tabular}{llccc}
\toprule
Model & Type & Blind & Descriptive & $\Delta$ \\
\midrule
Claude Opus 4.8    & Proprietary & $0.145 \pm 0.010$ & $0.561 \pm 0.007$ & $+0.417 \pm 0.008$ \\
GPT-5.5            & Proprietary & $0.096 \pm 0.008$ & $0.469 \pm 0.016$ & $+0.373 \pm 0.024$ \\
Claude Sonnet 4.6  & Proprietary & $0.118 \pm 0.010$ & $0.401 \pm 0.016$ & $+0.283 \pm 0.022$ \\
\midrule
Qwen3.5-122B-A10B  & Open (122B MoE) & $0.062 \pm 0.002$ & $0.282 \pm 0.018$ & $+0.219 \pm 0.018$ \\
Qwen3-Next-80B     & Open (80B MoE) & $0.122 \pm 0.011$ & $0.326 \pm 0.017$ & $+0.204 \pm 0.027$ \\
Llama 3.3-70B      & Open (70B) & $0.080 \pm 0.018$ & $0.258 \pm 0.015$ & $+0.178 \pm 0.029$ \\
Llama 4 Maverick   & Open ($\sim$400B MoE) & $0.085 \pm 0.015$ & $0.241 \pm 0.008$ & $+0.155 \pm 0.021$ \\
Gemma 3-12B        & Open (12B) & $0.066 \pm 0.009$ & $0.195 \pm 0.004$ & $+0.129 \pm 0.010$ \\
\bottomrule
\end{tabular}
\end{table*}

All eight models score higher with descriptive metadata on the full benchmark (\Cref{tab:main316}).
Blind accuracy ranges from 6.2\% to 14.5\%; descriptive accuracy ranges from 19.5\% to 56.1\%.
The gains range from 12.9 percentage points for Gemma~3-12B to 41.7 for Opus~4.8.
Supplying delivery information improves answers across the tested models, but even the best descriptive result leaves substantial room for improvement.

Per-type results appear in \Cref{tab:pertype}.
Affective-preference has the largest gap for both Opus and GPT, whereas affect-update has the smallest.
The ordering of the remaining types differs, so we do not infer a model-independent difficulty hierarchy.

\subsection{Which Metadata Helps?}
\label{sec:ablation}

\begin{table}[t]
\caption{Metadata and prompting ablations on all 316 questions ($n_d=5$, seed 42). All conditions include the transcript; wrong-metadata adds random emotion labels. CoT denotes chain-of-thought prompting. These single-seed runs are separate from the three-seed results in \Cref{tab:main316}.}
\label{tab:ablation316}
\centering
\small
\begin{tabular}{lrr}
\toprule
Condition & Opus 4.8 & GPT-5.5 \\
\midrule
blind              & 0.158 & 0.101 \\
wrong-metadata     & 0.215 & 0.146 \\
cot-blind          & 0.282 & 0.203 \\
events-only        & 0.304 & 0.212 \\
prosody-only       & 0.354 & 0.259 \\
emotion-only       & 0.513 & 0.427 \\
descriptive        & 0.576 & 0.478 \\
tagged             & 0.677 & 0.573 \\
cot-descriptive    & 0.766 & 0.709 \\
\bottomrule
\end{tabular}
\end{table}

Delivery descriptions outperform prosody, vocal events, and emotion labels alone (\Cref{tab:ablation316}). Structured tags improve overall accuracy further. Descriptions with chain-of-thought prompting~\citep{wei2022chain} give the highest accuracy for both models; transcript-only prompting helps but still trails descriptive input.

The comparison with emotion labels depends on the question set. Labels lead on direct questions but trail descriptions on indirect questions. Structured tags also match or trail descriptions on the indirect subset. A single emotion category may omit delivery details needed to interpret an action, preference, or stance. Task composition also differs between the subsets, so these results do not isolate phrasing as the cause.

Random emotion labels also score above the transcript-only baseline. A gain from adding annotations therefore does not by itself demonstrate use of the intended delivery information.

% \subsection{Distractor Scaling}
% \label{sec:distractor}

% With 3, 5, or 10 distractor sessions, descriptive input retains a positive gap for Opus, GPT, and Qwen3.5 (seed 42; \Cref{tab:distractor}).
% The gaps range from $+0.173$ to $+0.406$ across these runs.
% This limited, single-seed comparison does not establish a general context-length trend.

\subsection{Audio-native Ablations}
\label{sec:audio_results}

We evaluate Qwen2-Audio~\citep{chu2024qwen2audio} and Audio Flamingo~3~\citep{ghosh2026audio} on all 316 questions. Direct questions use synthesized windows around relevant turns; indirect questions use conversation clips. Each condition uses one greedy-decoding run and a lenient Claude Opus~4.5 judge that credits answers reaching the gold conclusion. Scores are therefore not directly comparable with \Cref{tab:main316}. More details in Appendix.

\begin{table}[t]
\caption{Audio ablations on 202 direct and 114 indirect questions. Metadata supplies textual delivery descriptions; the hint directs attention to delivery. Overall scores ($\approx$) are weighted estimates from rounded subgroup accuracies, with weights 202/316 and 114/316.}
\label{tab:audio316}
\centering
\small
\begin{tabular}{lrrr}
\toprule
Condition & Direct & Indirect & All \\
\midrule
\multicolumn{4}{l}{\textit{Qwen2-Audio}} \\
Audio only & 0.28 & 0.368 & $\approx 0.312$ \\
+ Hint & 0.32 & 0.412 & $\approx 0.353$ \\
+ Metadata & 0.20 & 0.491 & $\approx 0.305$ \\
+ Metadata + Hint & 0.24 & 0.535 & $\approx 0.346$ \\
\midrule
\multicolumn{4}{l}{\textit{Audio Flamingo 3}} \\
Audio only & 0.20 & 0.544 & $\approx 0.324$ \\
+ Hint & 0.24 & 0.518 & $\approx 0.340$ \\
+ Metadata & 0.16 & 0.614 & $\approx 0.324$ \\
+ Metadata + Hint & 0.16 & 0.614 & $\approx 0.324$ \\
\bottomrule
\end{tabular}
\end{table}

% Delivery descriptions help indirect questions but hurt direct questions for both models (\Cref{tab:audio316}); the opposing effects largely offset in the overall scores. The hint improves Qwen2-Audio throughout, while its effect on Audio Flamingo~3 is mixed. Audio with a hint gives the highest estimated overall accuracy for both models. Task composition and clip construction differ between the subsets, limiting conclusions about phrasing. These experiments test interpretation of short clips; recall from full audio histories remains untested.
 
\section{Conclusion}
\label{sec:conclusion}

VLME measures how access to vocal delivery changes answers about earlier conversations. Its main result is a consistent gain from delivery descriptions across eight models on a core selected to resist a transcript-only adversary. 

\section*{Limitations}
\label{sec:discussion}

The evidence dialogues and questions are authored, and the audio is synthesized. Their emotional associations and delivery patterns may differ from natural conversation, and the benchmark's intended answers are not a general account of a speaker's internal state. Explicit emotion labels supply an interpretation that acoustic descriptions do not; success with labels should not be read as success at perceiving emotion from speech.

The main experiments evaluate reasoning over supplied evidence in roughly 10--15k-token contexts. They do not test the full 100k-token histories or a deployed system's storage and retrieval policy. Adversarial filtering reduces transcript shortcuts for the designated model and prompt but does not certify universal unanswerability. The 321 derived questions were not separately filtered, and their answers are partly recoverable from transcripts under stronger prompting.

Scoring relies on LLM judges. Cross-judge and scrambled-context checks address some failure modes but do not replace human assessment. The audio study uses two 7B models on 104 synthesis outputs, whereas its text baselines use 114 items. Only 91 of those 104 clips passed the reported listening check. These sample and quality limits preclude broad claims about audio-native versus cascaded systems.

% \section*{Ethics Statement}
% The benchmark uses authored dialogues and synthetic speech; we collected no private user conversations or recordings. Audio synthesis uses acted RAVDESS recordings as conditioning references, so the generated clips should be distinguished from those source recordings. Acoustic descriptions document delivery, while emotion and pragmatic labels encode the benchmark's intended interpretation. Neither should be treated as a reliable diagnosis of a person's internal state. Models that retain delivery information may support intrusive profiling; deployments need user consent and controls over retention and deletion. We plan to release data and code under an open research license for academic and non-commercial evaluation.

% \section*{AI Use and Reproducibility}
% Generative AI was used for data generation, adversarial checks, scoring, manuscript editing, and evaluation-code assistance, but not for primary conceptualization or hypothesis formulation. We plan to release the data, scripts, and evaluation pipeline upon acceptance. \Cref{app:construction,app:prompts} document construction details and inference prompts.

\bibliography{mybib_v2}

@misc{maya1_model,
  author = {{Maya Research}},
  title = {Maya-1},
  year = {2025},
  howpublished = {Hugging Face model repository},
  url = {https://huggingface.co/maya-research/maya1}
}

@article{qwen2technicalreport,
  title={Qwen2 Technical Report},
  author={Yang, An and Baichuan, Chang and Bi, Junyang and Fang, Chiamin and Guo, Hongyi and Gao, He and Huang, Fei and Hui, Binyuan and Kong, Lingqiu and Kening, Lu and others},
  journal={arXiv preprint arXiv:2407.10671},
  year={2024}
}

@article{ghosh2026audio,
  title={Audio flamingo 3: Advancing audio intelligence with fully open large audio language models},
  author={Ghosh, Sreyan and Goel, Arushi and Kim, Jaehyeon and Kumar, Sonal and Kong, Zhifeng and Lee, Sang-gil and Yang, Chao-Han and Duraiswami, Ramani and Manocha, Dinesh and Valle, Rafael and others},
  journal={Advances in Neural Information Processing Systems},
  volume={38},
  pages={41819--41886},
  year={2026}
}

@inproceedings{xu2021beyond,
  title={Beyond Goldfish Memory: Long-Term Open-Domain Conversation},
  author={Xu, Jing and Szlam, Arthur and Weston, Jason},
  booktitle={Proceedings of the 60th Annual Meeting of the Association for Computational Linguistics (Volume 1: Long Papers)},
  pages={5180--5197},
  address={Dublin, Ireland},
  publisher={Association for Computational Linguistics},
  doi={10.18653/v1/2022.acl-long.356},
  year={2022}
}

@inproceedings{maharana2024locomo,
  title={Evaluating Very Long-Term Conversational Memory of {LLM} Agents},
  author={Maharana, Adyasha and Lee, Dong-Ho and Tulyakov, Sergey and Bansal, Mohit and Barbieri, Francesco and Fang, Yuwei},
  booktitle={Proceedings of the 62nd Annual Meeting of the Association for Computational Linguistics (Volume 1: Long Papers)},
  pages={13851--13870},
  address={Bangkok, Thailand},
  publisher={Association for Computational Linguistics},
  doi={10.18653/v1/2024.acl-long.747},
  year={2024}
}

@inproceedings{wu2025longmemeval,
  title={{LongMemEval}: Benchmarking Chat Assistants on Long-Term Interactive Memory},
  author={Wu, Di and Wang, Hongwei and Yu, Wenhao and Zhang, Yuwei and Chang, Kai-Wei and Yu, Dong},
  booktitle={International Conference on Learning Representations (ICLR)},
  year={2025},
  note={arXiv:2410.10813}
}

@article{wu2026longmemeval,
  title={{LongMemEval-V2}: Evaluating Long-Term Agent Memory Toward Experienced Colleagues},
  author={Wu, Di and Ji, Zixiang and Kawatkar, Asmi and Kwan, Bryan and Gu, Jia-Chen and Peng, Nanyun and Chang, Kai-Wei},
  journal={arXiv preprint arXiv:2605.12493},
  year={2026}
}

@article{jiang2025personamem,
  title={{PersonaMem-v2}: Towards Personalized Intelligence via Learning Implicit User Personas and Agentic Memory},
  author={Jiang, Bowen and Yuan, Yuan and Shen, Maohao and Hao, Zhuoqun and Xu, Zhangchen and Chen, Zichen and Liu, Ziyi and Vijjini, Anvesh Rao and He, Jiashu and Yu, Hanchao and Poovendran, Radha and Wornell, Gregory and Ungar, Lyle and Roth, Dan and Chen, Sihao and Taylor, Camillo Jose},
  journal={arXiv preprint arXiv:2512.06688},
  year={2025}
}

@article{wen2026amber,
  title={{A-MBER}: Affective Memory Benchmark for Emotion Recognition},
  author={Wen, Deliang and Sun, Ke and Wang, Yu},
  journal={arXiv preprint arXiv:2604.07017},
  year={2026}
}

@inproceedings{du2024perltqa,
  title={{PerLTQA}: A Personal Long-Term Memory Dataset for Memory Classification, Retrieval, and Fusion in Question Answering},
  author={Du, Yiming and Wang, Hongru and Zhao, Zhengyi and Liang, Bin and Wang, Baojun and Zhong, Wanjun and Wang, Zezhong and Wong, Kam-Fai},
  booktitle={Proceedings of the 10th SIGHAN Workshop on Chinese Language Processing (SIGHAN-10)},
  pages={152--164},
  address={Bangkok, Thailand},
  publisher={Association for Computational Linguistics},
  year={2024}
}

@article{kim2024dialsim,
  title={{DialSim}: A Dialogue Simulator for Evaluating Long-Term Multi-Party Dialogue Understanding of Conversational Agents},
  author={Kim, Jiho and Chay, Woosog and Hwang, Hyeonji and Kyung, Daeun and Chung, Hyunseung and Cho, Eunbyeol and Kwon, Yeonsu and Jo, Yohan and Choi, Edward},
  journal={arXiv preprint arXiv:2406.13144},
  year={2024}
}

@inproceedings{tan2025membench,
  title={{MemBench}: Towards More Comprehensive Evaluation on the Memory of {LLM}-based Agents},
  author={Tan, Haoran and Zhang, Zeyu and Ma, Chen and Chen, Xu and Dai, Quanyu and Dong, Zhenhua},
  booktitle={Findings of the Association for Computational Linguistics: ACL 2025},
  pages={19336--19352},
  address={Vienna, Austria},
  publisher={Association for Computational Linguistics},
  doi={10.18653/v1/2025.findings-acl.989},
  year={2025}
}

@inproceedings{zhao2025prefeval,
  title={Do {LLM}s Recognize Your Preferences? Evaluating Personalized Preference Following in {LLM}s},
  author={Zhao, Siyan and Hong, Mingyi and Liu, Yang and Hazarika, Devamanyu and Lin, Kaixiang},
  booktitle={International Conference on Learning Representations (ICLR)},
  year={2025},
  note={arXiv:2502.09597}
}

@inproceedings{huang2024dynamicsuperb,
  title={Dynamic-{SUPERB}: Towards a Dynamic, Collaborative, and Comprehensive Instruction-Tuning Benchmark for Speech},
  author={Huang, Chien-yu and Lu, Ke-Han and Wang, Shih-Heng and Hsiao, Chi-Yuan and Kuan, Chun-Yi and Wu, Haibin and Arora, Siddhant and Chang, Kai-Wei and Shi, Jiatong and Peng, Yifan and Sharma, Roshan and Watanabe, Shinji and Ramakrishnan, Bhiksha and Shehata, Shady and Lee, Hung-yi},
  booktitle={IEEE International Conference on Acoustics, Speech and Signal Processing (ICASSP)},
  year={2024}
}

@inproceedings{yang2024airbench,
  title={{AIR-Bench}: Benchmarking Large Audio-Language Models via Generative Comprehension},
  author={Yang, Qian and Xu, Jin and Liu, Wenrui and Chu, Yunfei and Jiang, Ziyue and Zhou, Xiaohuan and Leng, Yichong and Lv, Yuanjun and Zhao, Zhou and Zhou, Chang and Zhou, Jingren},
  booktitle={Proceedings of the 62nd Annual Meeting of the Association for Computational Linguistics (Volume 1: Long Papers)},
  pages={1979--1998},
  year={2024}
}

@inproceedings{wang2025audiobench,
  title={{AudioBench}: A Universal Benchmark for Audio Large Language Models},
  author={Wang, Bin and Zou, Xunlong and Lin, Geyu and Sun, Shuo and Liu, Zhuohan and Zhang, Wenyu and Liu, Zhengyuan and Aw, AiTi and Chen, Nancy F.},
  booktitle={Proceedings of the 2025 Conference of the Nations of the Americas Chapter of the Association for Computational Linguistics: Human Language Technologies (Volume 1: Long Papers)},
  pages={4297--4316},
  year={2025}
}

@inproceedings{sakshi2025mmau,
  title={{MMAU}: A Massive Multi-Task Audio Understanding and Reasoning Benchmark},
  author={Sakshi, S and Tyagi, Utkarsh and Kumar, Sonal and Seth, Ashish and Selvakumar, Ramaneswaran and Nieto, Oriol and Duraiswami, Ramani and Ghosh, Sreyan and Manocha, Dinesh},
  booktitle={The Thirteenth International Conference on Learning Representations (ICLR)},
  year={2025}
}

@inproceedings{wang2025mmsu,
  title={{MMSU}: A Massive Multi-task Spoken Language Understanding and Reasoning Benchmark},
  author={Wang, Dingdong and Li, Junan and Wu, Jincenzi and Yang, Dongchao and Chen, Xueyuan and Zhang, Tianhua and Meng, Helen},
  booktitle={The Fourteenth International Conference on Learning Representations (ICLR)},
  year={2026},
  note={arXiv:2506.04779}
}

@inproceedings{chen2025paras2s,
  title={{ParaS2S}: Benchmarking and Aligning Spoken Language Models for Paralinguistic-aware Speech-to-Speech Interaction},
  author={Yang, Shu-wen and Tu, Ming and Liu, Andy T. and Qu, Xinghua and Lee, Hung-yi and Lu, Lu and Wang, Yuxuan and Wu, Yonghui},
  booktitle={The Fourteenth International Conference on Learning Representations (ICLR)},
  year={2026},
  note={arXiv:2511.08723}
}

@inproceedings{ao2024sdeval,
  title={{SD-Eval}: A Benchmark Dataset for Spoken Dialogue Understanding Beyond Words},
  author={Ao, Junyi and Wang, Yuancheng and Tian, Xiaohai and Chen, Dekun and Zhang, Jun and Lu, Lu and Wang, Yuxuan and Li, Haizhou and Wu, Zhizheng},
  booktitle={Advances in Neural Information Processing Systems (NeurIPS) Datasets and Benchmarks Track},
  year={2024},
  eprint={2406.13340},
  archivePrefix={arXiv}
}

@inproceedings{cpbench2025,
  title={Benchmarking Contextual and Paralinguistic Reasoning in Speech-{LLM}s: A Case Study with In-the-Wild Data},
  author={Wang, Qiongqiong and Sailor, Hardik Bhupendra and Liu, Tianchi and Zhang, Wenyu and Huzaifah, Muhammad and Lertcheva, Nattadaporn and Sun, Shuo and Chen, Nancy F. and Wu, Jinyang and Aw, AiTi},
  booktitle={Findings of the Association for Computational Linguistics: EMNLP 2025},
  pages={14133--14148},
  year={2025}
}

@inproceedings{jiang2025s2sarena,
  title={{S2S-Arena}: Evaluating Paralinguistic Instruction Following in Speech-to-Speech Models},
  author={Jiang, Feng and Lin, Zhiyu and Liu, Yiyang and Xue, Liumeng and Bu, Fan and Du, Yuhao and Chen, Xiangying and Wang, Benyou and Li, Haizhou},
  booktitle={Proceedings of the 64th Annual Meeting of the Association for Computational Linguistics (Volume 1: Long Papers)},
  year={2026},
  note={arXiv:2503.05085}
}

@article{busso2008iemocap,
  title={{IEMOCAP}: Interactive Emotional Dyadic Motion Capture Database},
  author={Busso, Carlos and Bulut, Murtaza and Lee, Chi-Chun and Kazemzadeh, Abe and Mower, Emily and Kim, Samuel and Chang, Jeannette N. and Lee, Sungbok and Narayanan, Shrikanth S.},
  journal={Language Resources and Evaluation},
  volume={42},
  number={4},
  pages={335--359},
  year={2008},
  doi={10.1007/s10579-008-9076-6}
}

@inproceedings{poria2019meld,
  title={{MELD}: A Multimodal Multi-Party Dataset for Emotion Recognition in Conversations},
  author={Poria, Soujanya and Hazarika, Devamanyu and Majumder, Navonil and Naik, Gautam and Cambria, Erik and Mihalcea, Rada},
  booktitle={Proceedings of the 57th Annual Meeting of the Association for Computational Linguistics},
  pages={527--536},
  address={Florence, Italy},
  publisher={Association for Computational Linguistics},
  doi={10.18653/v1/P19-1050},
  year={2019}
}

@inproceedings{xu2024secap,
  title={{SECap}: Speech Emotion Captioning with Large Language Model},
  author={Xu, Yaoxun and Chen, Hangting and Yu, Jianwei and Huang, Qiaochu and Wu, Zhiyong and Zhang, Shi-Xiong and Li, Guangzhi and Luo, Yi and Gu, Rongzhi},
  booktitle={Proceedings of the AAAI Conference on Artificial Intelligence},
  volume={38},
  pages={19323--19331},
  doi={10.1609/aaai.v38i17.29902},
  year={2024}
}

@inproceedings{lin2024paralinguistics,
  title={Paralinguistics-Enhanced Large Language Modeling of Spoken Dialogue},
  author={Lin, Guan-Ting and Shivakumar, Prashanth Gurunath and Gandhe, Ankur and Yang, Chao-Han Huck and Gu, Yile and Ghosh, Shalini and Stolcke, Andreas and Lee, Hung-yi and Bulyko, Ivan},
  booktitle={IEEE International Conference on Acoustics, Speech and Signal Processing (ICASSP)},
  pages={10316--10320},
  year={2024}
}

@article{livingstone2018ravdess,
  title={The Ryerson Audio-Visual Database of Emotional Speech and Song ({RAVDESS}): A dynamic, multimodal set of facial and vocal expressions in North American English},
  author={Livingstone, Steven R. and Russo, Frank A.},
  journal={PLoS ONE},
  volume={13},
  number={5},
  pages={e0196391},
  year={2018}
}

@article{chu2024qwen2audio,
  title={{Qwen2-Audio} Technical Report},
  author={Chu, Yunfei and Xu, Jin and Yang, Qian and Wei, Haojie and Wei, Xipin and Guo, Zhifang and Leng, Yichong and Lv, Yuanjun and He, Jinzheng and Lin, Junyang and Zhou, Chang and Zhou, Jingren},
  journal={arXiv preprint arXiv:2407.10759},
  year={2024}
}

@article{hurst2024gpt4o,
  title={{GPT-4o} System Card},
  author={Hurst, Aaron and Lerer, Adam and Goucher, Adam P. and others},
  journal={arXiv preprint arXiv:2410.21276},
  year={2024}
}

@article{xu2025qwen25omni,
  title={{Qwen2.5-Omni} Technical Report},
  author={Xu, Jin and Guo, Zhifang and He, Jinzheng and Hu, Hangrui and He, Ting and Bai, Shuai and Chen, Keqin and Wang, Jialin and Fan, Yang and Dang, Kai and Zhang, Bin and Wang, Xiong and Chu, Yunfei and Lin, Junyang},
  journal={arXiv preprint arXiv:2503.20215},
  year={2025}
}

@article{defossez2024moshi,
  title={Moshi: A Speech-Text Foundation Model for Real-Time Dialogue},
  author={D{\'e}fossez, Alexandre and Mazar{\'e}, Laurent and Orsini, Manu and Royer, Am{\'e}lie and P{\'e}rez, Patrick and J{\'e}gou, Herv{\'e} and Grave, Edouard and Zeghidour, Neil},
  journal={arXiv preprint arXiv:2410.00037},
  year={2024}
}

@article{zeng2024glm4voice,
  title={{GLM-4-Voice}: Towards Intelligent and Human-Like End-to-End Spoken Chatbot},
  author={Zeng, Aohan and Du, Zhengxiao and Liu, Mingdao and Wang, Kedong and Jiang, Shengmin and Zhao, Lei and Dong, Yuxiao and Tang, Jie},
  journal={arXiv preprint arXiv:2412.02612},
  year={2024}
}

@misc{dia2025,
  author={{Nari Labs}},
  title={Dia: A 1.6B-parameter Dialogue Text-to-Speech Model},
  year={2025},
  publisher={Hugging Face / GitHub},
  howpublished={\url{https://huggingface.co/nari-labs/Dia-1.6B}}
}

@article{qwen3tts2026,
  title={{Qwen3-TTS} Technical Report},
  author={Hangrui Hu and Xinfa Zhu and Ting He and Dake Guo and Bin Zhang and Xiong Wang and Zhifang Guo and Ziyue Jiang and Hongkun Hao and Zishan Guo and Xinyu Zhang and Pei Zhang and Baosong Yang and Jin Xu and Jingren Zhou and Junyang Lin},
  journal={arXiv preprint arXiv:2601.15621},
  year={2026}
}

@article{pahwa2026audio2tool,
  title={Audio2Tool: Speak, Call, Act--A Dataset for Benchmarking Speech Tool Use},
  author={Pahwa, Ramit and Beedu, Apoorva and Priye, Parivesh and Gandhi, Rutu and Takawale, Saloni and Baijal, Aruna and Yang, Zengli},
  journal={arXiv preprint arXiv:2604.22821},
  year={2026}
}

@article{mcnemar1947note,
  title={Note on the sampling error of the difference between correlated proportions or percentages},
  author={McNemar, Quinn},
  journal={Psychometrika},
  volume={12},
  number={2},
  pages={153--157},
  year={1947},
  publisher={Cambridge University Press \& Assessment}
}

@book{efron2000introduction,
  title={An introduction to the bootstrap},
  author={Efron, Bradley and Tibshirani, Robert J and others},
  year={2000},
  publisher={Boca Raton, Florida}
}

@article{wei2022chain,
  title={Chain-of-thought prompting elicits reasoning in large language models},
  author={Wei, Jason and Wang, Xuezhi and Schuurmans, Dale and Bosma, Maarten and Xia, Fei and Chi, Ed and Le, Quoc V and Zhou, Denny and others},
  journal={Advances in neural information processing systems},
  volume={35},
  pages={24824--24837},
  year={2022}
}

@article{xie2026voicemem,
  title={VoiceMem: Streaming Dual-Brain Memory for Real-Time Interaction},
  author={Xie, Zhifei and Lang, Jiaqi and An, Ze and Zhao, Yifan and Yang, Dongchao and Li, Kai and Ma, Ziyang and Lin, Mingbao and Miao, Chunyan and Yan, Shuicheng},
  journal={arXiv preprint arXiv:2608.26005},
  year={2026}
}

@inproceedings{anand2026parallm,
  title={{ParA-LLM}: A Unified Approach to Paralinguistic and Acoustic Speech Understanding},
  author={Anand, Nishit and Su, Jiaqi and Chen, Ke and Wang, Yunyun and Manocha, Dinesh and Duraiswami, Ramani and Kumar, Rithesh and Jin, Zeyu},
  booktitle={Proceedings of Interspeech 2026},
  year={2026},
  eprint={2609.22771},
  archivePrefix={arXiv},
  url={https://arxiv.org/abs/2609.22771}
}

@inproceedings{pang2026voxparadox,
  title={Do Audio {LLM}s Listen or Read? Analyzing and Mitigating Paralinguistic Failures with {VoxParadox}},
  author={Pang, Jiacheng and Chaubey, Ashutosh and Soleymani, Mohammad},
  booktitle={Proceedings of the 43rd International Conference on Machine Learning},
  series={Proceedings of Machine Learning Research},
  volume={306},
  pages={96307--96334},
  publisher={PMLR},
  year={2026},
  url={https://proceedings.mlr.press/v306/pang26b.html}
}
\clearpage
\appendix
\section{Construction Details}
\label{app:construction}

\paragraph{Annotation schema.}
The 12 emotion labels are neutral, happy, excited, content, sad, disappointed, anxious, frustrated, angry, embarrassed, bored, and affectionate. Prosody fields cover rate, pitch, loudness, pauses, and emphasized words, which must occur verbatim in the turn. Voice events are laughs, sighs, coughs, clears\_throat, and gasps. Pragmatic flags encode sarcasm and uncertainty. A lexical filter excludes emotion terms, inflections, and roughly 60 interpretive glosses from free-text delivery descriptions. The blind transcript is byte-identical across input views; descriptive input adds acoustic stage directions, and tagged input exposes structured annotations.

\paragraph{Authoring and audit.}
Question-relevant conversations were authored in 14 themed batches (two pilots and twelve 16-item batches). Batches use distinct topics and persona names; collision scans check lexical overlap between instances. Sarcasm is capped at one item per batch. Identified shortcuts included pragmatic priors, outcome cues, gold answers matching event priors, assumed recovery, and explicit A/B alternatives. These informed subsequent authoring checks. On assembled histories, ranking sessions by emotive-annotation density retrieved the relevant session at top-1 in 13.1\% of cases (random: 3.3\%), below the specified 15\% threshold; a session-length probe scored zero. These probes address particular shortcuts rather than proving that all shortcuts are absent.

\paragraph{Long-history construction.}
For a question requiring $k$ relevant conversations, the full history adds 40 topic-screened LongMemEval filler sessions with neutral annotations and $6+2(k-1)$ emotionally expressive distractor sessions from a separate pool. Relevant conversations are placed at early, middle, or late positions; questions spanning multiple sessions retain their temporal ordering. These details describe the provided full histories. The reported model comparison instead uses the relevant conversations and five unrelated sessions (\Cref{sec:setup}).

\paragraph{Derived questions.}
The final indirect set contains 114 questions that reuse evidence sessions from the direct core. They ask about actions, preferences, stances, or intentions without explicitly referring to vocal delivery. Their task distribution is reported in \Cref{tab:types}; intent attribution appears only in this set. Evidence reuse does not establish that full histories, including distractors and ordering, are identical. Transcript-only checks cover every question in the final set (\Cref{sec:audit}). The earlier Indirect~v1 and Indirect~v2 experiment versions are identified separately in \Cref{sec:setup}.

\subsection{Speech Synthesis and Emotion Control}
\label{app:audio_synthesis}

\paragraph{Reference audio.}
Dia~\citep{dia2025} conditions speech on both text and reference audio. In our primary synthesis, the authored transcript supplies the words, while an acted RAVDESS recording~\citep{livingstone2018ravdess} matching the target emotion provides an acoustic example of the desired delivery. This allows delivery to vary without adding emotion names to the spoken dialogue. Dia also supports speaker markers \texttt{[S1]} and \texttt{[S2]} and nonverbal cues such as \texttt{(laughs)} and \texttt{(sighs)}. Reference conditioning guides expression but does not guarantee that every annotated cue is realized.

\paragraph{Text-based controls.}
The supplementary generators offer different ways to request delivery. The 1.7B CustomVoice and VoiceDesign variants of Qwen3-TTS~\citep{qwen3tts2026} accept natural-language instructions for emotion, tone, and speaking rate. Maya-1~\citep{maya1_model} combines a voice description with inline controls such as \texttt{<sarcastic>}, \texttt{<whisper>}, and \texttt{<sigh>}. An illustrative instruction is ``speak quietly and slowly, with long pauses''; a Maya utterance can include \texttt{<sigh>} before the spoken words. These examples explain the documented interfaces, rather than specifying the release's exact prompts.

\paragraph{Human review.}
Three human reviewers assessed all clips. Where review identified poor audio quality, we supplemented Dia generation with Qwen3-TTS and Maya-1. The target emotion is a generation condition, not a measurement of the generated speech.

\paragraph{Earlier synthesis and quality checks.}
The earlier Dia (1.6B) study used trimmed, peak-normalized RAVDESS references through a fixed 12-to-8 emotion mapping. The reference transcript was the first \texttt{[S1]} line; dialogue alternated \texttt{[S1]}/\texttt{[S2]} over a typically six-turn window, beginning on an assistant turn where possible. Settings were guidance scale 3.0, temperature 1.8, top-$p$ 0.9, and top-$k$ 45. The manifest recorded parameters, seeds, and references. Four speaker voices were used. An initial Whisper-large-v3-based emotion check reached about 30\% on acted RAVDESS and was made advisory. One listener then recorded pass/fail decisions: 91 of 104 clips passed. These quality figures refer to the earlier 104-clip study.
% TODO: Supply generator models, the 7B audit model, judge configuration/rubrics, decoding settings, the 12-to-8 mapping, Indirect v2 construction, reference handling, and clip-level quality/exclusion records.

\section{Error Analysis}
\label{sec:errors_appendix}

We group all 202 core items by correctness with and without delivery descriptions (\Cref{tab:errors}). Both-correct cases show that passing the designated audit does not prevent other models from answering from transcripts. Both-wrong cases identify remaining difficulty even with supplied descriptions; neither category alone establishes a capability ceiling or a leakage mechanism.

\begin{table*}[t]
\caption{Opus error analysis on Original 202Q.}
\label{tab:errors}
\centering
\small
\begin{tabular}{lrl}
\toprule
Outcome & $N$ & Description \\
\midrule
Descriptions help & 85 & Descriptive correct, blind wrong \\
Both wrong & 78 & Incorrect with either input \\
Both correct & 29 & Correct with either input \\
Descriptions hurt & 10 & Descriptive wrong, blind correct \\
\bottomrule
\end{tabular}
\end{table*}

\section{Inference Prompts}
\label{app:prompts}

\paragraph{Standard text prompt.}
\begin{quote}
\small
You are a personal assistant with access to the user's conversation history.

=== CONVERSATION HISTORY ===\\
\{history\}\\
=== END HISTORY ===

Today's date is \{question\_date\}. Answer the user's question based on the history above. Be concise (1--3 sentences). If the history does not contain enough information to answer, say so instead of guessing.

Question: \{question\}
\end{quote}

\paragraph{Delivery hint.}
The following instruction is prepended to the standard answer instruction. Its wording is retained for reproducibility; ``true feelings'' is the prompt's phrasing, not a claim about the reliability of emotion inference.
\begin{quote}
\small
When answering, consider not just \emph{what} was said but \emph{how} it was said. Delivery cues like tone, pace, volume, pauses, laughter, and sighs often reveal the speaker's true feelings, intentions, and preferences.
\end{quote}

\paragraph{Audio prompt.}
\begin{quote}
\small
Listen to this conversation. Answer concisely (1--3 sentences).

Question: \{question\}
\end{quote}

\section{Additional Results}
\label{app:results}

\subsection{Results by Question Type}

\begin{table*}[t]
\caption{Per-type affect gaps on 316 Questions($n_d=5$, three-seed mean $\pm$ standard deviation). Types are sorted by Opus gap.}
\label{tab:pertype}
\centering
\small
\begin{tabular}{lccc}
\toprule
Type & Opus $\Delta$ & GPT $\Delta$ & $N$ \\
\midrule
affective-preference     & $+0.613 \pm 0.023$ & $+0.560 \pm 0.069$ & 25 \\
prosody-disambiguated    & $+0.546 \pm 0.016$ & $+0.407 \pm 0.042$ & 36 \\
temporal-affective       & $+0.449 \pm 0.044$ & $+0.449 \pm 0.080$ & 26 \\
affect-recall            & $+0.398 \pm 0.042$ & $+0.343 \pm 0.070$ & 36 \\
cross-session-affect     & $+0.347 \pm 0.046$ & $+0.373 \pm 0.046$ & 25 \\
affect-update            & $+0.284 \pm 0.021$ & $+0.185 \pm 0.037$ & 27 \\
\bottomrule
\end{tabular}
\end{table*}

The affective-preference and affect-update types have the largest and smallest gaps, respectively, for both models.
Between those extremes, Opus and GPT differ in their ordering: GPT's cross-session-affect gap exceeds its affect-recall gap, whereas Opus shows the reverse.
The table excludes abstention variants.

\subsection{Question Sets and Hinting}

\begin{table*}[t]
\caption{Indirect 114 Question results with and without a retrieval-time hint. No hint and + Hint use descriptive input; Blind is the unprompted transcript-only baseline. Values are three-seed mean $\pm$ standard deviation.}
\label{tab:hint}
\centering
\small
\begin{tabular}{lcccc}
\toprule
Model & Blind & No hint & + Hint & Lift \\
\midrule
Qwen3.5-122B      & $0.066{\pm 0.008}$ & $0.148{\pm 0.027}$ & $0.571{\pm 0.008}$ & $+0.423{\pm 0.020}$ \\
Sonnet 4.6        & $0.129{\pm 0.013}$ & $0.259{\pm 0.011}$ & $0.591{\pm 0.034}$ & $+0.331{\pm 0.044}$ \\
Opus 4.8          & $0.169{\pm 0.011}$ & $0.305{\pm 0.016}$ & $0.631{\pm 0.009}$ & $+0.326{\pm 0.008}$ \\
GPT-5.5           & $0.143{\pm 0.014}$ & $0.285{\pm 0.025}$ & $0.562{\pm 0.018}$ & $+0.277{\pm 0.015}$ \\
Llama 3.3-70B     & $0.074{\pm 0.015}$ & $0.152{\pm 0.018}$ & $0.424{\pm 0.005}$ & $+0.271{\pm 0.019}$ \\
Qwen3-Next-80B    & $0.164{\pm 0.012}$ & $0.245{\pm 0.004}$ & $0.514{\pm 0.033}$ & $+0.269{\pm 0.030}$ \\
Gemma 3-12B       & $0.150{\pm 0.019}$ & $0.176{\pm 0.015}$ & $0.445{\pm 0.027}$ & $+0.269{\pm 0.022}$ \\
Llama 4 Maverick  & $0.126{\pm 0.011}$ & $0.207{\pm 0.012}$ & $0.286{\pm 0.007}$ & $+0.079{\pm 0.019}$ \\
\bottomrule
\end{tabular}
\end{table*}

The nuanced questions explicitly refer to voice, tone, or delivery; the core questions ask about affect; indirect questions use open-ended wording without a delivery cue.
These are separate item sets, rather than alternative phrasings of a matched set.
In the hinted condition, the instruction directs the model to consider delivery (\Cref{app:prompts}).
% TODO: The original spectrum table had four rows, not five. It is retained in unverified_results.tex pending run verification.
% TODO: Indirect gaps in tab:spectrum (e.g., Opus +0.179) do not equal the descriptive-minus-blind means in tab:hint (Opus 0.305-0.169). Preserve the source values pending run-level verification.

\subsection{Transcript-Only Controls}
\label{app:results_controls}

On Indirect~v1, hint-on-blind improves Opus by $+0.388$ (0.169 to $0.557\pm0.031$) and GPT by $+0.393$ (0.143 to $0.536\pm0.026$).
The resulting descriptive-minus-blind gaps under the hint are reported as $+0.074$ and $+0.026$.
On Original 202Q (Opus, seed 42), blind input improves by $+0.079$ (0.188 to 0.267), while descriptive input with the hint reaches 0.733.
The reported prompted gap is $+0.465$, compared with $+0.376$ without the hint.
% TODO: 0.733-0.267 is 0.466 after rounding; confirm the +0.465 gap against unrounded paired results. The original also reports +0.367 for indirect hint lift, inconsistent with +0.388 elsewhere; do not use that value without verification.

Two controls examine the hinted indirect results.
With scrambled evidence sessions, accuracy is 0.043.
Re-judging the hint-on-blind outputs with GPT-5.5 gives 0.600, compared with 0.536 under the Opus~4.5 judge.

These checks support dependence on the conversation and robustness to this judge substitution. They do not show that prompting recovers an acoustic signal absent from the input.
% TODO: Confirm matched model, item set, and seeds for the +0.067 comparison. The 0.631 descriptive+hint value is attributed to Opus in tab:hint, while the preceding 0.536 blind+hint value is attributed to GPT in the prose. The source does not identify the model for scrambled context or wrong-metadata+hint. Do not present +0.067 as a verified causal estimate until run IDs are checked.

\subsection{Distractor Count}

\begin{table*}[t]
\caption{Accuracy with 3, 5, or 10 distractor sessions (Original 316Q, seed 42). Each run retains a positive descriptive--blind gap.}
\label{tab:distractor}
\centering
\small
\begin{tabular}{llccc}
\toprule
Model & $n_d$ & Blind & Descriptive & $\Delta$ \\
\midrule
Opus 4.8 & 3  & 0.188 & 0.594 & $+0.406$ \\
         & 5  & 0.193 & 0.564 & $+0.371$ \\
         & 10 & 0.158 & 0.525 & $+0.366$ \\
\midrule
GPT-5.5  & 3  & 0.144 & 0.441 & $+0.297$ \\
         & 5  & 0.144 & 0.475 & $+0.332$ \\
         & 10 & 0.134 & 0.446 & $+0.312$ \\
\midrule
Qwen3.5  & 3  & 0.114 & 0.287 & $+0.173$ \\
         & 5  & 0.089 & 0.302 & $+0.213$ \\
         & 10 & 0.099 & 0.272 & $+0.173$ \\
\bottomrule
\end{tabular}
\end{table*}

Descriptive accuracy remains above blind accuracy in every listed run.
The variation is not monotonic: GPT and Qwen3.5 improve from 3 to 5 distractors before decreasing at 10, whereas Opus decreases across the three settings.
Single-seed results cannot establish a general context-length trend.

\end{document}